\documentclass[sigconf]{acmart}
\AtBeginDocument{%
  }

\acmConference[SIGSPATIAL '26]{The 34th ACM International Conference on
  Advances in Geographic Information Systems}{November 3--6, 2026}{Riverside, CA, USA}
\acmYear{2026}
\copyrightyear{2026}
\usepackage{booktabs}
\usepackage{tabularx}
\usepackage{textcomp}
\usepackage{subcaption}
\usepackage[]{todonotes}

\setcopyright{acmlicensed}
\acmDOI{XXXXXXX.XXXXXXX}
\acmConference[Conference acronym 'XX]{Make sure to enter the correct
  conference title from your rights confirmation email}{June 03--05,
  2018}{Woodstock, NY}
\acmISBN{978-1-4503-XXXX-X/2018/06}
\begin{document}

\title{Reconstructing GRACE Terrestrial Water Storage with Spatio-Temporal Graph Neural Networks: An Application to South America [Applications]}


\author{Lukas Arzoumanidis}
\email{lukas.arzoumanidis@hcu-hamburg.de}
\orcid{https://orcid.org/0000-0001-6668-1695}
\affiliation{%
  \institution{HafenCity University, \\ Computational Methods Lab}
  \city{Hamburg}
  \country{Germany}
}

\author{Lara Johannsen, Klara Middendorf}
\email{\{firstname.lastname\}@hcu-hamburg.de}
\orcid{0009-0008-8503-6868}
\affiliation{%
  \institution{HafenCity University, \\ Geodesy and Adjustment Theory}
  \city{Hamburg}
  \country{Germany}
}


\author{Annette Eicker}
\email{annette.eicker@gfz.de}
\orcid{0000-0002-9087-1445}
\affiliation{
  \institution{GFZ Helmholtz Centre for Geosciences, \\ Global Geomonitoring and Gravity Field}
  \city{Potsdam}
  \country{Germany}
}

\author{Youness Dehbi}
\email{youness.dehbi@hcu-hamburg.de}
\orcid{https://orcid.org/0000-0003-0133-4099}
\affiliation{%
  \institution{HafenCity University, \\ Computational Methods Lab}
  \city{Hamburg}
  \country{Germany}
}
\renewcommand{\shortauthors}{Arzoumanidis et al.}

\begin{abstract}
Terrestrial water storage (TWS) integrates snow, soil moisture, surface water,
and groundwater and is a key indicator of how climate variability and human
activity reshape the global water cycle. The GRACE and GRACE-FO satellite
missions provide the only direct, globally consistent observations of TWS
change, but their record only begins in 2002 which is too short for many
climate-scale analyses. We present a deep learning application that
reconstructs monthly GRACE-like TWS anomalies (TWSA) back to 1940 by learning
the relationship between daily ERA5 meteorological forcing (precipitation,
evapotranspiration, runoff) and monthly GRACE observations. In contrast to
prior reconstruction approaches based on grid-cell-wise regression, CNNs, or LSTMs, we
adapt a multivariate time series graph neural network (MTGNN) architecture, which was originally developed for mobility and traffic forecasting on urban sensor networks to this satellite-geodesy task.
Spatial dependencies are encoded in a
static, interpretable hybrid adjacency matrix that combines geodesic proximity
with lagged correlations of climatic time series, capturing both local
hydrological coupling and large-scale teleconnections. Evaluated over South
America against GRACE/GRACE-FO (2002--2023), the reconstruction achieves a
grid-cell Pearson correlation of 0.69, a basin-mean correlation of 0.94, and a near-zero bias, and it
reproduces the spatial fingerprints of the 2015/16 El Ni\~no and 2020/21
La Ni\~na events. A systematic comparison with established reconstruction approaches
(GTWS-MLrec, RM-REC, GRAiCE) shows that the graph-based model is statistically
competitive at basin scale, reaching a correlation within 0.025 of the best
baseline while using only roughly half to a tenth of the predictors the other models require and revealing characteristic weaknesses in arid regions in all models.
We discuss best practices and lessons learned from
deploying graph deep learning in a satellite-geodesy application, and outline
extensions via additional predictors and physics-informed constraints based on
the terrestrial water balance equation. To support reproducibility and future research, the complete implementation is publicly available at \url{https://github.com/hcu-cml/MTGNN-TWS-Reconstruction-GRACE}.
\end{abstract}

\ccsdesc[500]{Applied computing~Earth and atmospheric sciences}
\ccsdesc[300]{Computing methodologies~Supervised learning by regression}
\ccsdesc[500]{Computing methodologies~Neural networks}

\keywords{Terrestrial water storage, GRACE, Graph neural networks,
Spatio-temporal modeling, ERA5, Climate reconstruction, GeoAI}

\maketitle

\section{Introduction}
Understanding changes in the Earth's water storage is crucial for quantifying
the impacts of climate variability and anthropogenic activity on the global
water cycle. Terrestrial water storage (TWS) comprises the water stored above
and below the land surface in the form of snow, ice, soil moisture, surface
water, and groundwater, and acts as a key integrator of hydrological
processes. TWS is highly sensitive to climate change: it registers shifts in
precipitation, temperature, and extreme weather, and its variations relate
directly to the occurrence of droughts, floods, and changes in seasonal water
availability. Accurate TWS data are therefore essential for separating natural
variability from human-induced trends and for managing freshwater resources
under accelerating global change \citep{yin2023,humphrey2019}.

Since 2002, the Gravity Recovery and Climate Experiment \citep{tapey_2004} and its
successor GRACE-FO \citep{landerer_2020} have provided the only direct, globally consistent
measurements of TWS change. By tracking variations in the distance
between two co-orbiting satellites, the missions resolve monthly changes in
the Earth's gravity field, from which mass redistribution can be
inferred, including continental water movement, ice mass change, and
groundwater depletion \citep{jensen2020,tapley2019}. Unlike sparse
ground-based monitoring networks, GRACE captures subsurface and large-scale
hydrological processes with homogeneous accuracy, making it a benchmark
dataset of Earth system science, particularly in regions where in-situ
networks are weak or absent \citep{jensen2020}.

The observational record, however, spans barely more than two decades. This is
a severe limitation for climate science, where multi-decadal time series are
required to identify trends, detect tipping points, and attribute variability
to large-scale drivers such as the El Ni\~no--Southern Oscillation (ENSO)
\citep{humphrey2019}. To cover the pre-GRACE era, a growing body of work therefore
reconstructs GRACE-like signals from meteorological data using statistical and
machine learning methods (Section~\ref{sec:related}). Yet most existing
approaches, which are grid-cell-wise regressions, CNNs, and LSTMs, treat grid
cells as spatially independent or only implicitly model spatial context. They
thereby ignore the inherently networked character of the water cycle, in which
hydrological connectivity and atmospheric teleconnections couple distant
regions.

This paper reports on the application of spatio-temporal \emph{graph} deep
learning to this reconstruction problem. Notably, the architecture we deploy
was not designed for the Earth sciences at all: the multivariate time series
graph neural network (MTGNN) of \citet{wu2020} originates in the mobility
domain, where it was developed for forecasting traffic conditions on urban
road-sensor networks. The structural analogy that motivates the transfer is straightforward. 
In traffic forecasting, sensors form the nodes of a graph, congestion propagates along
the road network, and the model must capture how a disturbance at one
location influences readings elsewhere with a delay. In continental
hydrology, grid cells form the nodes, water and atmospheric moisture
propagate along river systems and circulation patterns, and a precipitation
anomaly in one region influences storage elsewhere with a lag of days to
months. Both are multivariate time series regression problems on a fixed set
of spatially embedded nodes whose mutual influence is structured, directional
in time, and only partially explained by geometric proximity. We adapt MTGNN
to the gravimetric setting where $1^\circ$ grid cells become graph nodes carrying
daily ERA5 forcing variables, and a static, interpretable adjacency
matrix, constructed from geodesic distance and lagged climatic
correlations, replaces both the road network and MTGNN's learned graph,
encoding local hydrological coupling as well as long-range teleconnections.
Trained on the GRACE/GRACE-FO period, the model reconstructs monthly TWS
anomalies (TWSA) back to 1940, the start of the ERA5 reanalysis. The implementation presented here focuses on South America (1{,}120 land nodes) as a test region within a globally designed workflow.

Our main contributions are practical and methodological:
\begin{itemize}
  \item We demonstrate that a graph architecture from urban mobility
        forecasting transfers to gravimetric satellite data with
        adaptation, and we document the design decisions, most notably
        replacing the learned graph structure with a static, domain-informed
        hybrid adjacency matrix, that made the application robust and
        interpretable.
  \item We provide a multi-stage evaluation against GRACE/GRACE-FO
        (2002--2023), covering quantitative metrics, seasonal and spatial
        error structure, interannual residuals, and the spatial fingerprints
        of the 2015/16 El Ni\~no and 2020/21 La Ni\~na events.
    \item We benchmark the reconstruction against three existing approaches
        spanning the methodological spectrum (GTWS-MLrec \citep{yin2023},
        RM-REC \citep{li2021}, GRAiCE \citep{palazzoli2025}) and show
        that a graph model is statistically competitive at basin scale 
        (correlation 0.94 with GRACE) \emph{using only three input
        variables}, where the better-scoring baselines consume from roughly
        six to over twenty predictors, while sharing the community-wide
        weakness in arid, human-influenced regions.
  \item We report lessons learned and two concrete extension paths. Input
        parameter expansion (a temperature pilot already raises grid-cell
        correlation from 0.69 to 0.71) and physics-informed training via the
        terrestrial water balance equation.
\end{itemize}

\section{Background and Problem Definition}
\label{sec:background}
This section provides the domain context needed to follow the application
without a background in geodesy or hydrology, and then states the learning
problem formally.

\subsection{Why Satellite Gravimetry with GRACE?}
GRACE, launched in 2002 as a collaboration between the National Aeronautics and Space Administration (NASA) and Deutsches Zentrum für Luft- und Raumfahrt (DLR) (eng. \emph{German Aerospace Center}), consists of two
satellites flying on the same orbit, one trailing the other. When the leading
satellite passes over a region of slightly higher mass, such as a mountain range or an
aquifer after a wet season, it is accelerated marginally earlier than its
follower, and the inter-satellite distance changes. A K-band
microwave ranging system measures these distance variations continuously. 
In addition to that GRACE-FO (launched 2018 after GRACE ended in 2017) adds a laser ranging
interferometer of even higher precision \citep{tapley2019,chen2022}. Over the
course of a month the satellite pair samples the entire globe, and the
accumulated ranging data are inverted into a monthly model of the Earth's
gravity field.

These monthly gravity fields are expressed as \emph{spherical harmonic
coefficients}---a frequency-domain representation of a function
on the sphere, analogous to a 2D Fourier transform, where low degrees encode
planetary-scale structures and higher degrees encode finer spatial details. The
solutions used here resolve up to degree 96 and after the necessary spatial filtering have an effective spatial resolution of roughly 300\,km) \citep{kurtenbach2012,chen2022}.
Subtracting a static mean field isolates the \emph{time-variable} component,
which over land is dominated by water mass redistribution \citep{sun2017}.
After standard corrections (Section~\ref{sec:gracedata}), the coefficients are transformed onto a geographic grid and expressed as \emph{equivalent water height} (EWH), representing the thickness of a hypothetical water layer that would produce the observed gravity change. A value of
$-0.2$\,m EWH at a grid cell thus means the cell has lost mass equivalent to
20\,cm of water spread over its area, regardless of whether that loss
occurred in soil moisture, surface water, or groundwater. Values are reported
as \emph{anomalies} (TWSA) relative to a 2004--2009 baseline mean, so the
quantity of interest is the deviation from a reference state, not absolute
storage. This integrated view is what makes GRACE unique, as groundwater depletion at depths of hundreds of meters is invisible to optical and radar satellites but detectable by gravimetry.

\subsection{Reanalysis Data}
The model input comes from ERA5 \citep{hersbach2020}, an atmospheric
\emph{reanalysis}. A reanalysis is best understood as a globally consistent,
physics-constrained interpolation of the historical observation record: a
numerical weather model is run over the past, and at every assimilation step
the model state is optimally corrected toward all available observations
(satellites, weather stations, radiosondes, ships, aircraft)
\citep{copernicus2023}. The result is a gap-free, gridded estimate of
atmospheric and land-surface variables that is dynamically consistent across
space, time, and variables, unlike raw station data, which are sparse and
heterogeneous. Crucial for this application, ERA5 extends back to 1940,
more than six decades before the first GRACE observation, with global hourly
coverage. Its quality is not uniform over time (fewer observations constrain
the early decades, and the satellite era begins around 1979), a caveat that
applies to every reconstruction built on it.

\subsection{Problem Definition}
\label{sec:problem}
Let $V = \{v_1, \dots, v_N\}$ be the set of $N = 1{,}120$ land grid cells of
the $1^\circ$ South American domain, embedded in a weighted undirected graph
$G = (V, E, A)$ with adjacency matrix $A \in \mathbb{R}^{N \times N}$
(Section~\ref{sec:graph}). For each calendar month $m$, the input is a window
of $T = 30$ daily data of $D = 3$ ERA5 flux variables
(precipitation, evapotranspiration, runoff) at every node,
$X_m \in \mathbb{R}^{T \times N \times D}$, and the target is the GRACE TWSA
of that month at every node, $y_m \in \mathbb{R}^{N}$. The task is to learn a
function
\begin{equation}
  f_\theta : \bigl(X_m,\, G\bigr) \longmapsto \hat{y}_m,
  \label{eq:problem}
\end{equation}
i.e., a node-level regression conditioned on the graph structure, by
supervised training on the months for which GRACE observations exist
(April 2002 -- December 2023). Reconstruction then consists of applying the
trained $f_\theta$ to the ERA5 record outside the supervision period
(1940--2002), where no gravimetric ground truth exists. Two properties make
this a spatial-computing problem rather than a generic regression: the
physical processes linking input and target are spatially networked (water
moves between cells; atmospheric teleconnections couple distant regions), and
the supervision signal lives at a coarser \emph{temporal} resolution
(monthly) than the input (daily), so the model must learn both spatial
aggregation over the graph and temporal aggregation over the window. The
implicit assumption underlying any such reconstruction, shared by all
related work in Section~\ref{sec:related}, is that the mapping from
meteorological forcing to storage response learned in the satellite era is
stationary enough to be applied to earlier decades.

\section{Related Work}
\label{sec:related}
Approaches for extending TWSA beyond the GRACE/GRACE-FO period fall into three
categories: statistical methods, physically based hydrological models, and,
increasingly, machine learning. We summarize the data-driven reconstructions
that serve as methodological context and, later, as comparison baselines.

\citet{humphrey2019} present GRACE-REC, a statistical reconstruction of
climate-driven TWS from 1901 to 2019. A linear reservoir model with seasonally
varying, temperature-parameterized residence time is updated daily from
precipitation and temperature, calibrated per GRACE mascon on de-trended and
de-seasonalized GRACE data. An ensemble of 100 simulations quantifies parameter and residual
uncertainty. Despite its simplicity, GRACE-REC matches or outperforms complex
hydrological models in reproducing interannual variability.

\citet{li2021} develop a global reconstruction (1979--2020, $0.5^\circ$;
referred to here as RM-REC) combining
machine learning (ANN, ARX, and MLR) with statistical mode decomposition and
time series decomposition. Rather than fitting each grid cell independently, it
reconstructs a few leading spatial modes of the GRACE field over regions of
varying size (continents, multi-basins, and basins), which allows it to
assimilate predictors located outside the study area, such as sea surface
temperature and climate indices, alongside precipitation, temperature, soil
moisture, runoff, and evaporation. Trained on 2002--2017 GRACE mascons and
validated against GRACE-FO, satellite laser ranging, and global mean sea level,
it reproduces strong El Ni\~no signals well.

\citet{yin2023} present GTWS-MLrec, a global reconstruction (1940--2022) from
an ensemble of five machine learning and statistical models trained pixel by
pixel with locally selected predictors from meteorological, hydrological,
land use, and vegetation categories. Per-cell model selection over eight input
schemes yields several consistent global products tied to different mascon
solutions.

\citet{palazzoli2025} introduce GRAiCE (1984--2021, $0.5^\circ$), training LSTM
and BiLSTM models per grid cell with lags of up to 24 months and Optuna-based
hyperparameter search. The five meteorological predictors are total
precipitation, snow depth water equivalent, surface net solar radiation,
surface air temperature, and relative humidity, with solar-induced
fluorescence added as an optional sixth predictor. The reconstructions achieve
global Pearson correlations above 0.9 against GRACE/GRACE-FO and capture ENSO
extremes well, but accuracy declines in arid, human-influenced regions.

\citet{gentner2025} propose DeepRec (1941--2023), a CNN encoder over
$17.5^\circ$ patches followed by an LSTM, fed by 16 ERA5 variables together
with sea surface temperature, the ENSO index, ISIMIP land-use and lake
fractions, and engineered geographic/temporal features, and targeting the
\emph{full} TWS signal including human-influenced trends, with validation
against global mean sea level and satellite laser ranging.

Table~\ref{tab:comparison} contrasts these approaches. Common to all is that
spatial structure is treated implicitly (per-cell models) or through regular
convolution on grids. Graph neural networks, in contrast, operate directly on
irregular neighborhood structures and have proven effective in climate and
geospatial flow applications, e.g., ENSO forecasting with graph convolutions
\citep{ruhlingcachay2020}, spatially explicit GeoAI on networks
\citep{sun2021,zhou2020}, place characterization \citep{ward2022}, or semantic segmentation of historical urban plans \citep{Arzoumanidis04032026}. 
Within SIGSPATIAL, GNNs have recently been applied to inter-county food flow
prediction \cite{food_flow_GNN_2025} and related spatial-network problems, underscoring their ability to generalize across spatial scales from topological regularities. To our
knowledge, this work is the first to apply spatio-temporal
graph deep learning to TWS reconstruction, explicitly encoding hydroclimatic
teleconnections in the graph topology.

\begin{table}[htbp]
  \centering
  \caption{Published GRACE reconstruction approaches used for context and comparison. Spatial resolution is given in degree-sized patches.}
  \label{tab:comparison}
  \small
  \begin{tabular}{l l l l}
    \toprule
    Model & Type & Period & Res. \\
    \midrule
    GRACE-REC \citep{humphrey2019} & Statistical memory  & 1901--2019 & $0.5^\circ$ $\times$ $0.5^\circ$ \\
    MTGNN (ours)                   & Spatio-temporal GNN & 1940--2023 & $1^\circ$ $\times$ $1^\circ$\\
    GRAiCE \citep{palazzoli2025}   & (Bi)LSTM            & 1984--2021 & $0.5^\circ$ $\times$ $0.5^\circ$ \\
    GTWS-MLrec \citep{yin2023}     & ML ensemble         & 1940--2022 & $0.25^\circ$ $\times$ $0.25^\circ$ \\
    DeepRec \citep{gentner2025}    & CNN + LSTM          & 1941--2023 & $0.5^\circ$ $\times$ $0.5^\circ$ \\
    RM-REC \citep{li2021}          & ML + decomposition  & 1979--2020 & $0.5^\circ$ $\times$ $0.5^\circ$\\
    \bottomrule
  \end{tabular}
\end{table}

\section{Data}
\label{sec:data}
The application links two physically related datasets through the terrestrial
water balance: GRACE/GRACE-FO observations of integrated storage change serve
as the prediction target, and ERA5 reanalysis fluxes serve as model input.

\subsection{GRACE/GRACE-FO Terrestrial Water Storage}
\label{sec:gracedata}
We use monthly gravity field solutions from the ITSG-Grace2018 series (GRACE)
and ITSG-Grace\_operational (GRACE-FO) provided by Graz University of
Technology \citep{mayerguerr2018}, processed with the GROOPS software
\citep{mayerguerr2021}. Standard corrections are applied: degree-1
(geocenter) coefficients are restored following \citet{cheng2013,sun2017};
the $C_{20}$ coefficient is replaced by satellite laser ranging estimates
\citep{chen2022}; glacial isostatic adjustment is removed with a GIA model;
and a DDK3 decorrelation filter suppresses striping errors and
high-frequency noise \citep{qian2022}. The corrected spherical harmonic
coefficients (up to degree 96) are converted to equivalent water heights
(EWH) \citep{guo2014} and expressed as anomalies relative to the 2004--2009
GRACE baseline. Gaps in the monthly record, including the 2017/18 inter-mission
gap, are interpolated, and a leakage mask removes coastal cells
affected by ocean signal leakage \citep{eicker2020}. The resulting target
dataset covers April 2002 to December 2023 on a global geographical
$1^\circ\times1^\circ$ grid with the South American subset, used in this work, containing
1{,}120 grid cells.

\subsection{ERA5 Meteorological Forcing}
ERA5, the fifth-generation atmospheric reanalysis of the European Centre for Medium-Range Weather Forecasts (ECMWF)\footnote{\url{https://www.ecmwf.int/}}, assimilates
decades of observations into a globally consistent estimate of the atmosphere
from 1940 to the present \citep{hersbach2020,copernicus2023}. We use three
single-level variables that constitute the terrestrial water balance
\citep{copernicus2023}, namely, total precipitation~($P$), surface latent heat flux
converted to evapotranspiration~($E$), and total runoff~($R$). Hourly fields
are aggregated to daily values, re-gridded to the same $1^\circ$ grid,
expressed in EWH via a spherical harmonic expansion to degree 96, and
provided as NetCDF ensuring full spatial consistency with the GRACE prediction target.

\subsection{Physical Link: Water Balance Equation}
The two datasets are physically coupled through the terrestrial water balance
\citep{lorenz2014,eicker2020}:
\begin{equation}
  P - E - R = \frac{dS}{dt},
  \label{eq:waterbalance}
\end{equation}
where the storage change $dS/dt$ is observed by GRACE and the flux terms are
represented by ERA5.

\section{Methodology}
\label{sec:method}
\subsection{Spatio-Temporal Graph-Based Architecture}
Conventional CNNs assume a regular grid with fixed local neighborhoods. This assumption is problematic for global climate data because important relationships are not always local. Regions separated by large distances may still exhibit strong interactions through river systems, ocean currents, and large-scale atmospheric circulation patterns \citep{wu2019,ruhlingcachay2020}. Graph convolution generalizes
the convolution operation to variable, unstructured neighborhoods and has
delivered strong results in climate applications such as ENSO forecasting
\citep{ruhlingcachay2020}. Spatio-temporal GNNs (STGNNs) combine per-time-step
graph convolution with temporal convolution or recurrence over node
sequences, processing input tensors $X \in \mathbb{R}^{T \times N \times D}$
($T$ time steps, $N$ nodes, $D$ features) \citep{wu2019}.

\subsection{MTGNN}
\label{sec:transfer}
The architecture we deploy, MTGNN \citep{wu2020}, was developed and
demonstrated in the mobility and traffic forecasting domain: predicting
future readings of road-sensor networks from their multivariate history,
where each sensor is a graph node and the (partly latent) road topology
governs how congestion propagates. We selected it for this application not
despite but \emph{because} of that origin. The traffic problem and the
hydrological problem share their abstract structure which can be defined as a fixed set of spatially
embedded nodes, multivariate time series per node, propagation of influence
along a network with characteristic delays, and the need to capture
periodicities at multiple time scales (rush hours and weekly cycles in
traffic; rainy seasons and annual cycles in hydrology). MTGNN's dilated
temporal convolutions, designed for exactly such multi-scale periodicity, and
its mix-hop graph convolutions, designed for delayed multi-node propagation,
therefore map naturally onto hydroclimatic dynamics.
Table~\ref{tab:transfer} summarizes the correspondence and the adaptations
that were required.

\begin{table}[htbp]
  \centering
  \caption{Transfer mapping from the original mobility setting of MTGNN
           \citep{wu2020} to the gravimetric application.}
  \label{tab:transfer}
  \small
  \begin{tabularx}{\linewidth}{@{}>{\raggedright\arraybackslash}p{3.8cm} X@{}}
    \toprule
    Mobility forecasting & TWSA reconstruction (this work) \\
    \midrule
    Road sensors as nodes & $1^\circ$ ERA5 grid cells as nodes ($N{=}1120$); \\
    Speed/volume readings & Daily $P$, $E$, $R$ fluxes ($D{=}3$); \\
    Congestion propagation along roads & Water propagation along river systems, atmospheric teleconnections; \\
    Learned latent adjacency & Static hybrid adjacency from geodesic distance and lagged climate correlations (Sec.~\ref{sec:graph}); \\
    Rush-hour/weekly periodicity & Seasonal/annual hydrological cycles; \\
    Multi-step forecasting of the same quantity & Cross-variable regression: monthly TWSA from a 30-day flux window; \\
    Abundant sensor history & 183 training months (short, gap-affected satellite record);\\
    \bottomrule
  \end{tabularx}
\end{table}

Three adaptations deserve emphasis. First, the prediction task changes from
\emph{forecasting} (predicting future values of the input quantity itself) to
\emph{cross-variable regression with temporal aggregation}. The model maps a
30-day window of daily fluxes to a single monthly value of a different
physical quantity, observed by a different instrument. Second, the graph is
prescribed rather than learned, for reasons detailed in
Section~\ref{sec:graph}. 
Third, the supervision
regime is far more constrained: instead of years of dense sensor data, only
183 monthly target fields exist for training, which pushed the design toward
strong regularization (dropout, weight decay) and a deliberately moderate
model capacity. 

\subsection{MTGNN Adapted to Hydrology}
\begin{figure}[htbp]
  \centering
  \includegraphics[width=\linewidth]{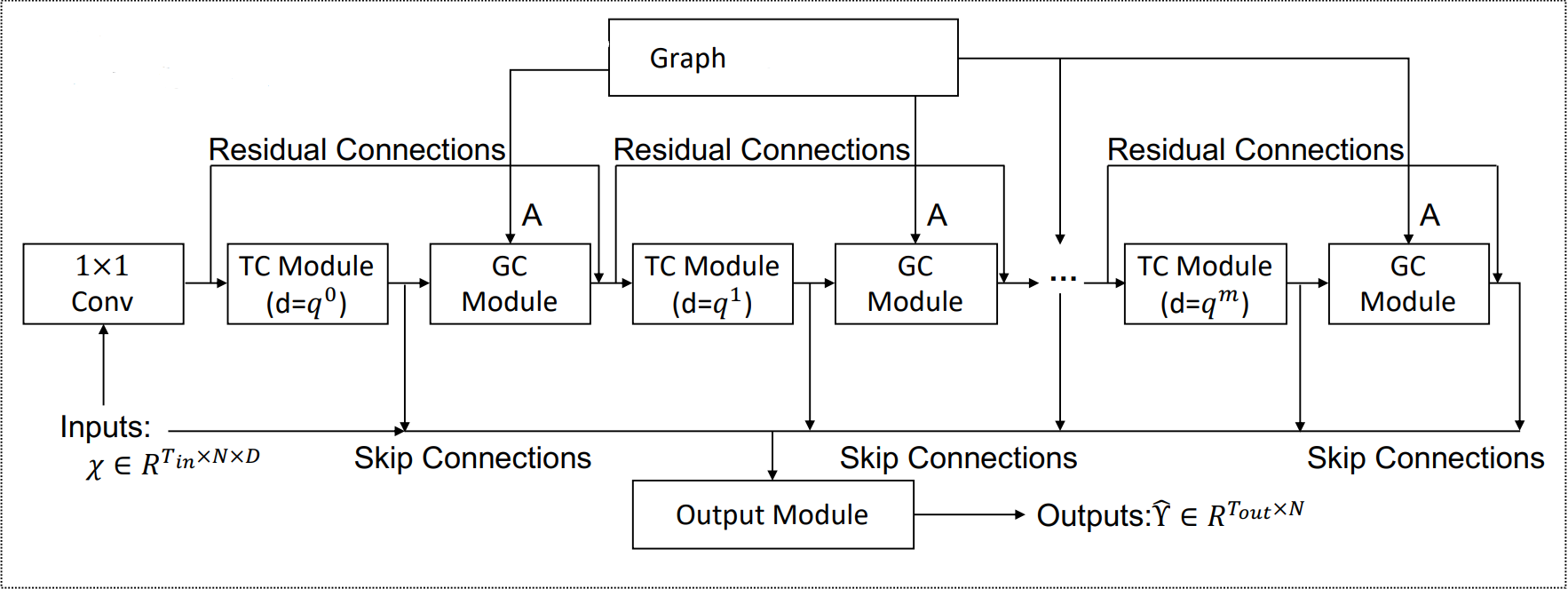}
  \caption{Model architecture of the adapted MTGNN (modified from \citet{wu2020}).}
    \Description{Enjoying the baseball game from the third-base
  seats. Ichiro Suzuki preparing to bat.}
  \label{fig:master}
\end{figure}

We retain the three functional components of MTGNN \citep{wu2020}: a graph
(structure) layer, graph convolution modules, and temporal convolution
modules. Graph convolution uses two mix-hop
propagation layers that aggregate information over multiple hop distances
while a retention factor counteracts over-smoothing \citep{wu2020,zhou2020}.
Temporal convolution uses dilated inception layers with kernel sizes
$1\times2$, $1\times3$, $1\times6$, and $1\times7$ and a tanh/sigmoid as activation functions for the gating
mechanism, yielding exponentially growing receptive fields that capture
delayed hydrological responses. Residual and skip connections stabilize
training and preserve node-level history \citep{wu2020}. Figure~\ref{fig:master}
shows the adapted architecture: nodes are $1^\circ$ ERA5 grid cells, node
features are daily $P$, $E$, $R$ sequences, and the output is one TWSA value
per node and month.

\subsection{A Static, Interpretable Hybrid Graph}
\label{sec:graph}
The original MTGNN \emph{learns} its adjacency matrix from randomly
initialized node embeddings with subgraph sampling \citep{wu2020}. For this
application we deliberately replaced the learned structure with a static
graph, for reasons that we consider an important practical lesson. First,
learned graphs are hard to interpret, whereas transparency about spatial
dependencies is essential in a geoscientific setting. Second, jointly
optimizing a latent graph and a deep predictive model proved prone to
convergence instability on heterogeneous meteorological inputs, while random
subgraph sampling can sever physically meaningful dependencies
\citep{huang2025,kirschstein2024}. 

\begin{figure}[htbp]
  \centering
  \includegraphics[width=\linewidth]{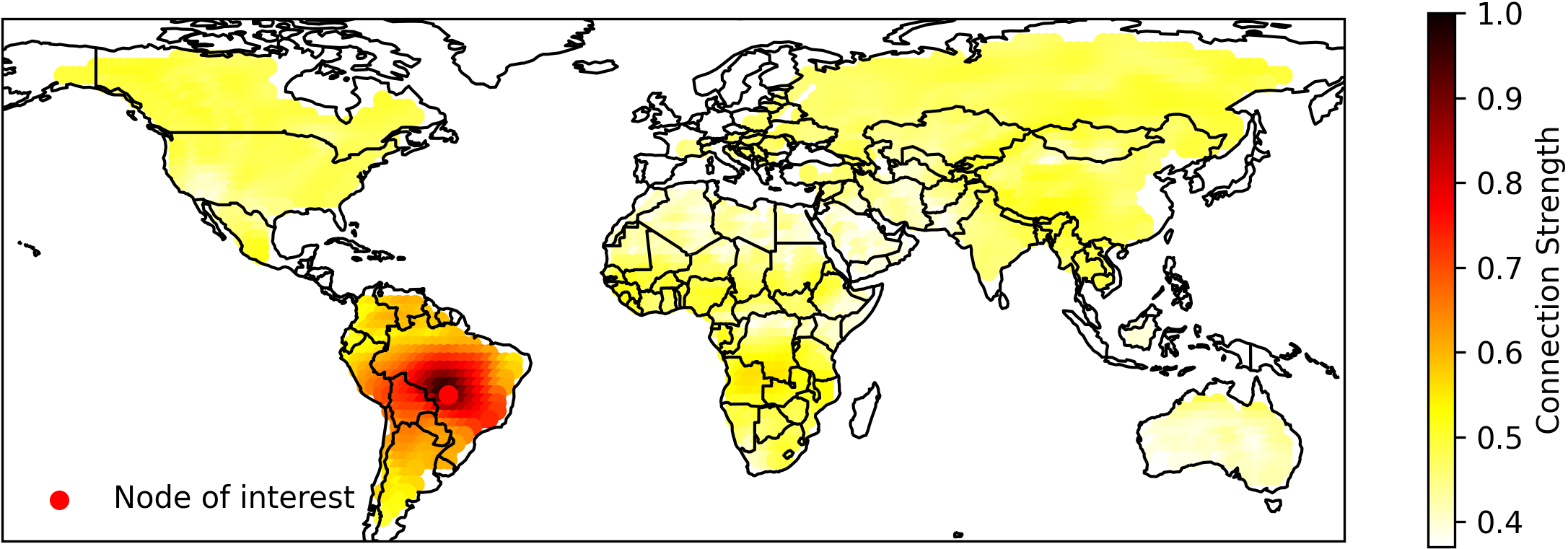}
  \caption{Adjacency weights of a single node in South America under the
           hybrid graph construction (Eq.~\ref{eq:lincomb}).}
             \Description{Enjoying the baseball game from the third-base
  seats. Ichiro Suzuki preparing to bat.}
  \label{fig:adjacency}
\end{figure}

The static graph is undirected and homogeneous, and its weights combine two
similarity terms. \emph{Climate similarity} captures teleconnections via the
maximum-lag Pearson correlation between node time series
\citep{silva2020,berman2016}. Here, $\rho^{(f)}_{ij}$ is the Pearson correlation coefficient, which measures linear dependence on a scale from $-1$ to $1$, between the time series of variable $f$ at nodes $i$ and $j$. It is evaluated at the lag $\tau \in [0,\tau_{\max}]$ that gives the largest absolute correlation. Using this lagged
cross-correlation rather than the zero-lag value captures relationships in
which one region's signal leads or trails another's, as is typical of
propagating or transport-driven teleconnection patterns. The per-variable
correlations are aggregated into a single similarity,
\begin{equation}
  s^{\mathrm{climate}}_{ij} = \sum_f w_f \cdot \frac{|\rho^{(f)}_{ij}| + 1}{2},
  \label{eq:climsim}
\end{equation}
where the absolute value lets strong anti-correlations contribute as much as
strong positive ones, and $w_f$ weights each variable.
\emph{Spatial proximity} converts the geodesic distance $d_{ij}$ into a
similarity with a distance-decay Gaussian kernel \citep{zhu2002}:
\begin{equation}
  s^{\mathrm{distance}}_{ij} = \exp\!\left(-\frac{d_{ij}}{\sigma}\right),
  \label{eq:dist}
\end{equation}
where the length scale $\sigma$ (in the units of $d_{ij}$) controls the rate of
distance decay. At $d_{ij}=\sigma$, the similarity falls to
$1/e \approx 0.37$ of its maximum value. Small $\sigma$ emphasizes local
connections, while large $\sigma$ permits stronger links between distant cells. 
The trade-off adjacency weight is their convex combination,
\begin{equation}
  A_{ij} = \alpha \cdot s^{\mathrm{climate}}_{ij}
         + (1 - \alpha) \cdot s^{\mathrm{distance}}_{ij},
  \label{eq:lincomb}
\end{equation}
with $\alpha = 0.6$ and $\sigma = 3500\,\mathrm{km}$ selected for the South American domain.
Figure~\ref{fig:adjacency} shows the resulting edge weights for a single node.
In addition to nearby locations, distant regions with similar climate conditions
can also receive substantial weight. These long-range connections capture
teleconnection patterns that grid-based models cannot represent.

\subsection{Training Setup}
\label{sec:training}
\begin{figure}[htbp]
  \centering
  \includegraphics[width=\linewidth]{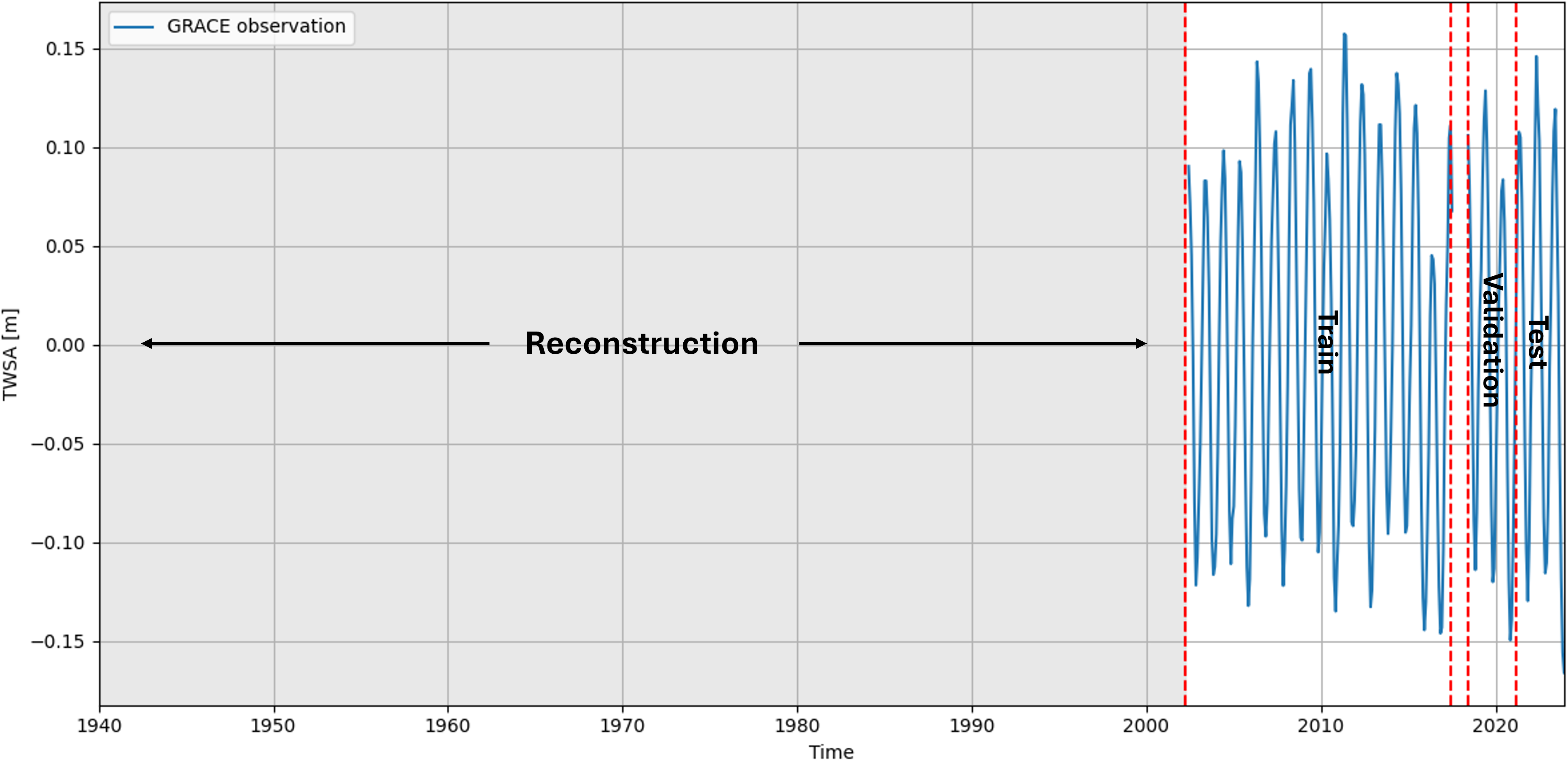}
  \caption{Sequential train/validation/test split along the GRACE record.}
  \label{fig:splitting}
    \Description{Enjoying the baseball game from the third-base
  seats. Ichiro Suzuki preparing to bat.}
\end{figure}

Each training example pairs a 30-day window of daily ERA5 inputs with the
GRACE TWSA of the corresponding month, i.e.,
$x \in \mathbb{R}^{30 \times 1120 \times 3} \rightarrow
 y \in \mathbb{R}^{1 \times 1120 \times 1}$.
Feature-wise Z-score normalization is fitted on the training partition only
and applied consistently to validation and test data \citep{singh2022}. 
Because random splits leak information in time series, the
data are split sequentially \citep{reitermanova2010}. Training covers April
2002 to June 2017 (183 months, containing the 2004--2009 GRACE baseline and
excluding the interpolated inter-mission gap), followed by 34 validation and
33 test months as can be seen in Figure~\ref{fig:splitting}.

The model is trained for 40 epochs with the Adam optimizer
\citep{kingma2015} (learning rate $10^{-3}$, batch size 64, dropout 0.4,
weight decay 0.005) and an MSE loss, which penalizes the large deviations
associated with hydrological extremes more strongly than MAE
\citep{duarte2021}. Architecture hyperparameters were tuned manually against
validation error resulting in a graph convolution depth of 4, 32 convolution channels, 64
residual and skip channels and 128 end channels. Training and validation losses
decrease monotonically without divergence, indicating no overfitting. The
final validation metrics are MSE 0.0117, RMSE 0.1083, MAE 0.0830 (normalized
units). On the held-out test months the model achieves MSE 0.0128,
RMSE 0.1131, and MAE 0.0862, which is consistent with validation and evidence of
stable generalization. Temporally aggregated test predictions track the
observed basin-mean signal closely, while the spatial correlation of
time-averaged fields is 0.61, foreshadowing the regional weaknesses analyzed
below.

\section{Experimental Results}
\label{sec:results}
The evaluation proceeds in three stages: (i) direct validation against
GRACE/GRACE-FO over 2002--2023, (ii) analysis of seasonal, spatial, and
interannual error structure including ENSO case studies, and (iii) a
systematic comparison with established reconstruction approaches.

\subsection{Correlation with GRACE Observations}
Table~\ref{tab:quality} summarizes the correlation between predictions and observations over South America from April 2002 to December 2023. All metrics are computed at the grid-cell level and pooled across all land cells and months. As a result, they measure performance at the native $1^\circ$ resolution rather than after spatial averaging. Given a grid-cell dynamic range of approximately $\pm 0.6\,\mathrm{m}$ EWH, the RMSE of $0.132\,\mathrm{m}$ and MAE of $0.096\,\mathrm{m}$ indicate high overall correlation. The bias is close to zero ($-0.004\,\mathrm{m}$), suggesting no systematic over- or underestimation. The grid-cell Pearson correlation of $0.69$ indicates a strong positive relationship between predicted and observed variability. This correlation is lower than the value obtained after spatial averaging. This difference is expected because spatial averaging reduces local errors and emphasizes large-scale seasonal variability. In contrast, the grid-cell metric preserves regional differences, including higher correlation in the humid Amazon and weaker correlation in the arid southern regions, as discussed in Sec.~\ref{sec:spatial}.

\begin{table}[htbp]
   \caption{Model evaluation against GRACE over South America (2002--2023).}
  \label{tab:quality}
  \begin{tabular}{lc}
    \toprule
    Quality measure & Value \\
    \midrule
    Root Mean Squared Error (RMSE) [m EWH] & 0.1323 \\
    Mean Absolute Error (MAE) [m EWH]      & 0.0955 \\
    Bias [m EWH]                           & $-0.0044$ \\
    Pearson correlation (all grid cells)   & 0.6931 \\
    \bottomrule
\end{tabular}
\end{table}

\begin{figure}[htbp]
  \centering
  \begin{subfigure}{\linewidth}
    \centering
    \includegraphics[width=\linewidth]{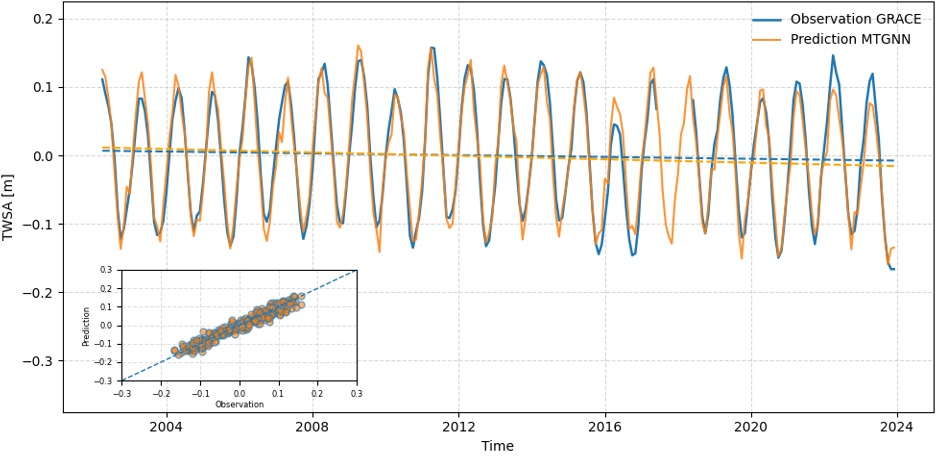}
    \caption{area mean over South America}
    \label{fig:ts-areamean}
  \end{subfigure}
  \\[1ex]
  \begin{subfigure}{\linewidth}
    \centering
    \includegraphics[width=\linewidth]{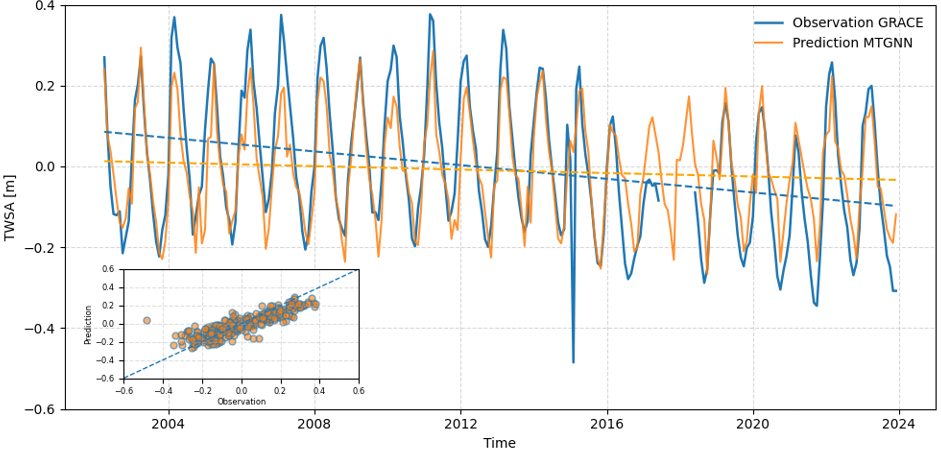}
    \caption{exemplary grid cell in the Amazon basin}
    \label{fig:ts-cell}
  \end{subfigure}
  \caption{Observed vs.\ predicted TWSA with scatter plots for (a) the area
           mean and (b) an exemplary Amazon grid cell (Figure~\ref{fig:adjacency}).}
  \label{fig:ts}
\end{figure}

Figure~\ref{fig:ts} compares predicted and observed time series for the area
mean and an exemplary Amazon grid cell (Figure~\ref{fig:adjacency}). The prediction follows the seasonal
cycle closely, and the scatter plots cluster around the identity line.
Smaller deviations concentrate in extreme wet and dry phases, and are slightly more pronounced at the single
cell than in the spatial mean, indicating lower consistency for local
phenomena. The exemplary cell also exhibits a clearly negative observed
long-term trend whose direction the model reproduces but whose magnitude it
underestimates.


\begin{figure}[htbp]
  \centering
  \includegraphics[width=0.6\linewidth]{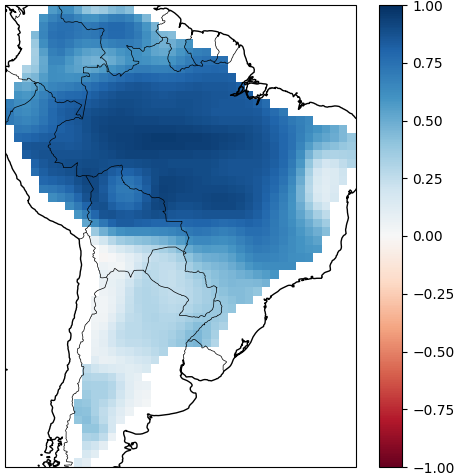}
  \caption{Per-grid-cell Pearson correlation between the MTGNN reconstruction
           and GRACE over South America (April 2002--December 2023). Correlation is
           highest in the central Amazon, where the seasonal storage signal is
           most pronounced, and falls below 0.5 in the arid south (parts of
           Bolivia, Paraguay, and Argentina, Chile), revealing the regional structure
           behind the aggregate grid-cell correlation reported in
           Table~\ref{tab:quality}.}
   \Description{Enjoying the baseball game from the third-base
  seats. Ichiro Suzuki preparing to bat.}         
  \label{fig:corrmap}
\end{figure}

The spatial correlation map (Figure~\ref{fig:corrmap}) shows where the model
is strongest. The central Amazon basin, where the seasonal storage signal is
most pronounced, reaches the highest correlations, confirming that the model
excels in regions dominated by strong, regular hydrological cycles.

\subsection{Seasonal and Spatial Error Structure}
\label{sec:spatial}
\begin{figure}[htbp]
  \centering
  \includegraphics[width=\linewidth]{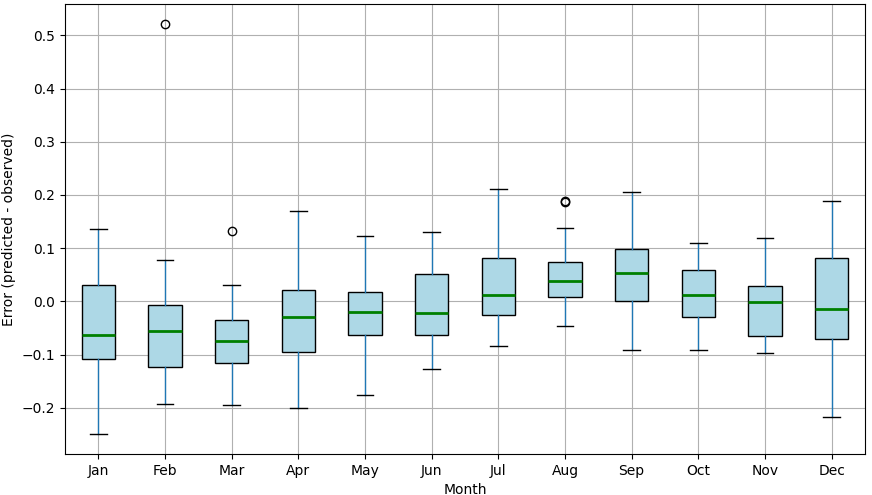}
  \caption{Monthly distribution of errors (predicted $-$ observed) at the
           exemplary Amazon grid cell.}
             \Description{Enjoying the baseball game from the third-base
  seats. Ichiro Suzuki preparing to bat.}
  \label{fig:boxplot}
\end{figure}

A monthly error analysis at the Amazon cell (Figure~\ref{fig:boxplot})
reveals systematic seasonal behavior. The TWSA is slightly underestimated from
December to June and slightly overestimated in the second half of the year,
with the largest error variability in the transition months between dry and
rainy season. The model thus struggles most at the onset and end of the rainy
season, and occasional outliers point to difficulties with extreme events.

\begin{figure}[htbp]
  \centering
  \includegraphics[width=\linewidth]{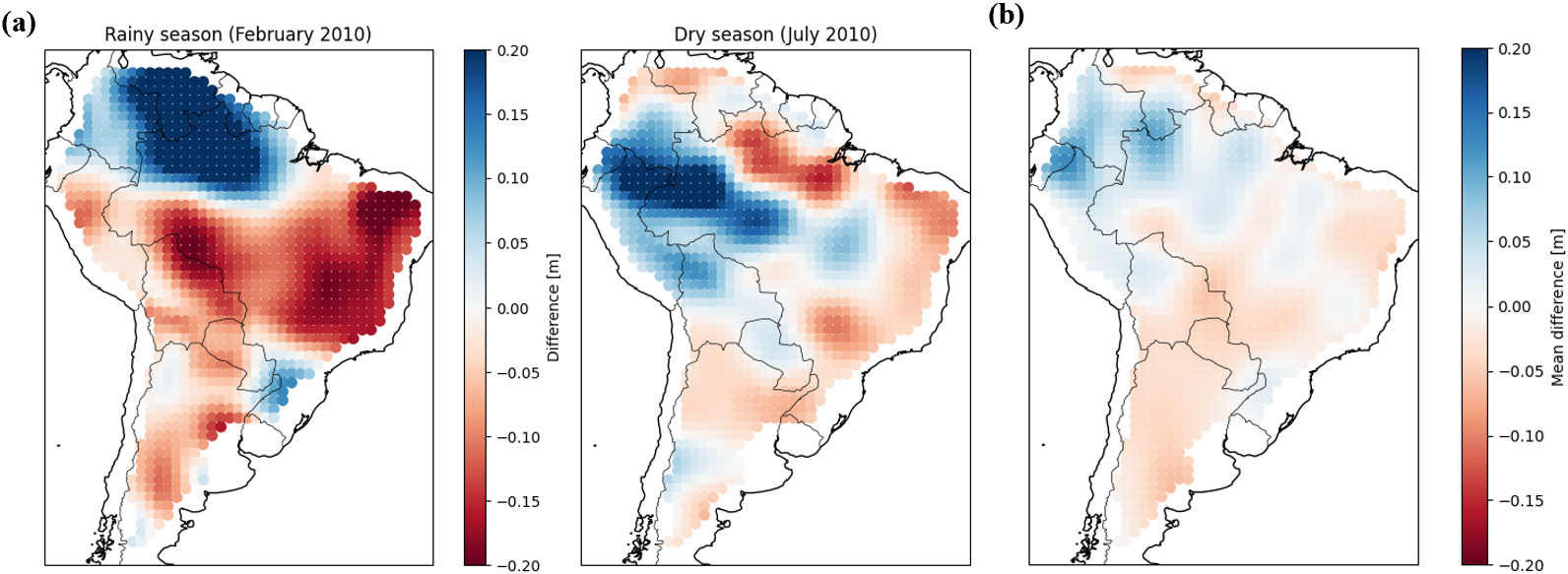}
  \caption{Spatial difference between prediction and observation for (a) rainy
           vs.\ dry season and (b) the full-period mean.}
             \Description{Enjoying the baseball game from the third-base
  seats. Ichiro Suzuki preparing to bat.}
  \label{fig:diffmaps}
\end{figure}

Seasonal difference maps (Figure~\ref{fig:diffmaps}) highlight the strong influence of spatial information on the error analysis.
Rainy season errors (e.g., February 2010) are large but
spatially coherent, whereas dry season errors (e.g., July 2010) are
fragmented, suggesting that the model's ability to \emph{distribute} water
within the region degrades at low storage levels. The long-term mean
difference exposes a persistent regional bias with a slight overestimation in the
northwestern basin and underestimation in the southeast.

\subsection{Interannual Signals and ENSO Case Study}
\begin{figure}[htbp]
  \centering
  \includegraphics[width=\linewidth]{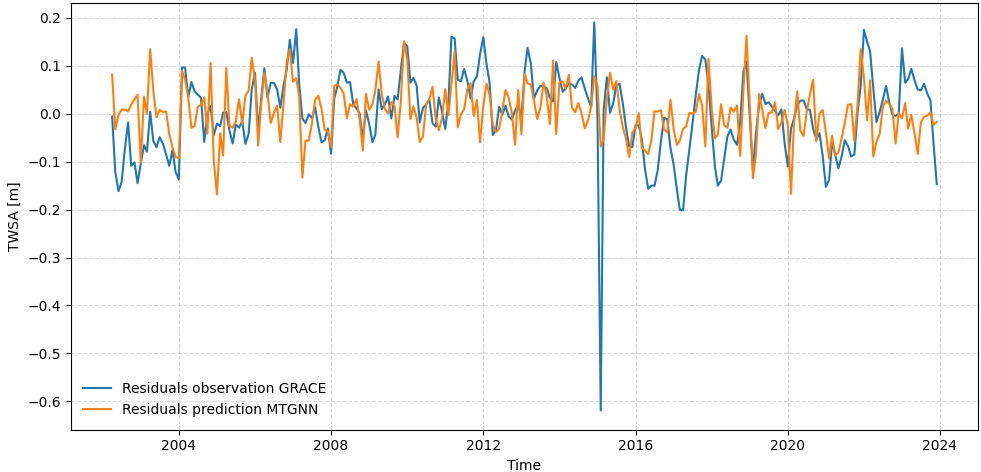}
  \caption{Residual (de-trended, de-seasonalized) TWSA of prediction and
           observation at exemplary Amazon grid cell. Note that the outlier in 2015 is known and results from poor sampling in a repeat orbit, meaning the error is due to low quality ground truth data rather than the proposed approach.}
             \Description{Enjoying the baseball game from the third-base
  seats. Ichiro Suzuki preparing to bat.}
  \label{fig:residuals}
\end{figure}

To test whether the model captures more than trend and seasonality, we fit a
harmonic regression with linear trend and (semi-)annual components
\citep{esa2020},
\begin{equation}
  \hat{S}(t) = a + b t + c\cos(\omega t) + d\sin(\omega t)
             + e\cos(2\omega t) + f\sin(2\omega t),
  \label{eq:lsfit}
\end{equation}
and analyze the residuals, which carry the interannual signal including
extreme events. At the Amazon cell (Figure~\ref{fig:residuals}), predicted and
observed residuals agree in waveform, timing, and magnitude. Additionally, peaks and troughs
coincide, multi-year excursions are followed, and amplitudes are comparable,
though the prediction is somewhat smoothed at extreme outliers.

\begin{figure}[htbp]
  \centering
  \includegraphics[width=\linewidth]{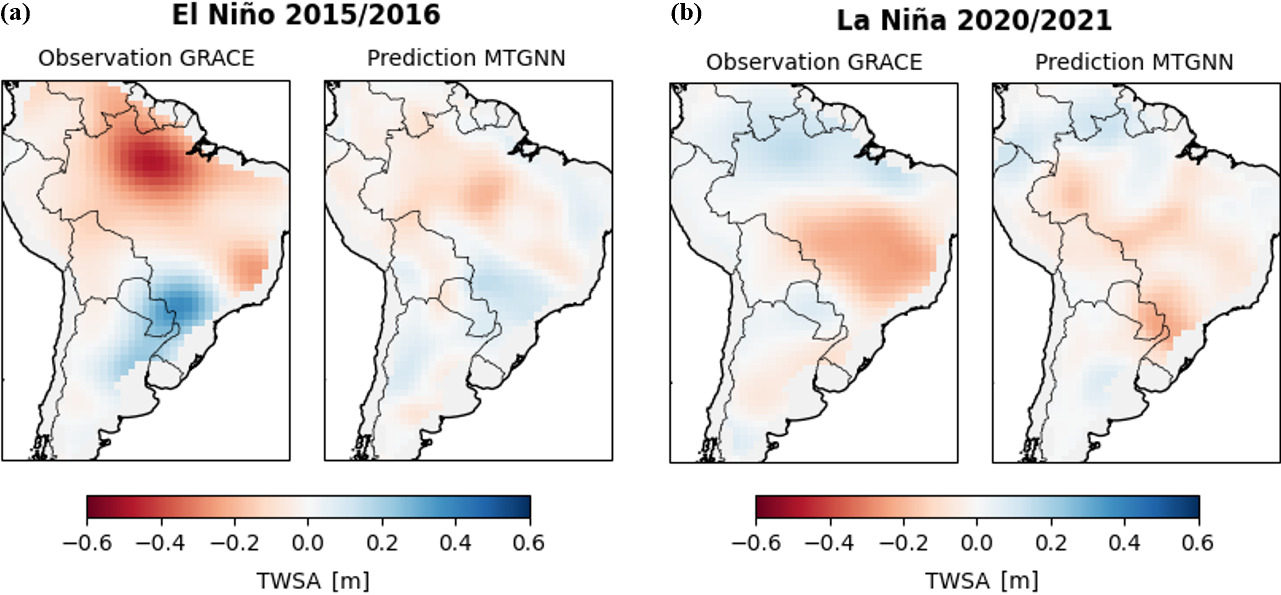}
  \caption{Deviations from the climatological monthly mean in observation and
           prediction during (a) the 2015/16 El Ni\~no and (b) the 2020/21
           La Ni\~na.}
           \Description{Enjoying the baseball game from the third-base
  seats. Ichiro Suzuki preparing to bat.}
  \label{fig:enso}
\end{figure}

Figure~\ref{fig:enso} examines the two strongest recent ENSO events as
spatial anomaly maps relative to the climatological monthly mean. During the
2015/16 El Ni\~no, GRACE shows large-scale drought (negative anomalies) in the
Amazon and pronounced wet anomalies in the La Plata basin. Our model
reproduces this contrasting pattern with correct extent and position, though
with slightly weaker intensity. During the 2020/21 La Ni\~na, the reversed
pattern, drought in southern Brazil, wet conditions in the Amazon, is
likewise captured. The model thus identifies the characteristic spatial
fingerprints of large-scale climate anomalies while underestimating their
amplitude, consistent with the smoothing seen in the residual analysis.

\subsection{Comparison with SOTA Reconstruction Approaches}
\label{sec:comparison}
We compare against GTWS-MLrec \citep{yin2023}, RM-REC \citep{li2021}, and
GRAiCE \citep{palazzoli2025}, which together span statistical, ensemble-ML,
and recurrent neural approaches. All products are harmonized to a common
$1^\circ$ grid, masked to the South American domain, and restricted to the
shared overlap period 2002--2020, so that methodological differences are not
confounded with technical ones. Table~\ref{tab:compmetrics} reports metrics on
the spatially averaged time series plus the correlation at the exemplary
Amazon cell.

\begin{table}[htbp]
    \caption{Comparison of established reconstruction approaches against GRACE based on basin-mean time series. \emph{Corr.\ (Amazon)} denotes the correlation at a single exemplary grid cell in the Amazon, while \emph{Corr.} refers to the basin-mean (spatially averaged) time series.} 
  \label{tab:compmetrics}
  \small
  \begin{tabular}{lccccc}
    \toprule
    Model & Inputs & Corr. & Std.\ dev. & RMSE & Corr.(Amazon)\\
    \midrule
    GRACE      & --       & 1.000 & 0.083 & --    & 1.000 \\
    MTGNN (ours)  & 3 & 0.939 & 0.088 & 0.030 & 0.866 \\
    GRAiCE     & $6$  & 0.964 & 0.083 & 0.026 & 0.907 \\
    GTWS-MLrec & $16$ & 0.940 & 0.074 & 0.031 & 0.932 \\
    RM-REC     & $25$  & 0.931 & 0.078 & 0.031 & 0.902 \\
    \bottomrule
\end{tabular}
\end{table}

\begin{figure}[htbp]
  \centering
  \includegraphics[width=0.95\linewidth]{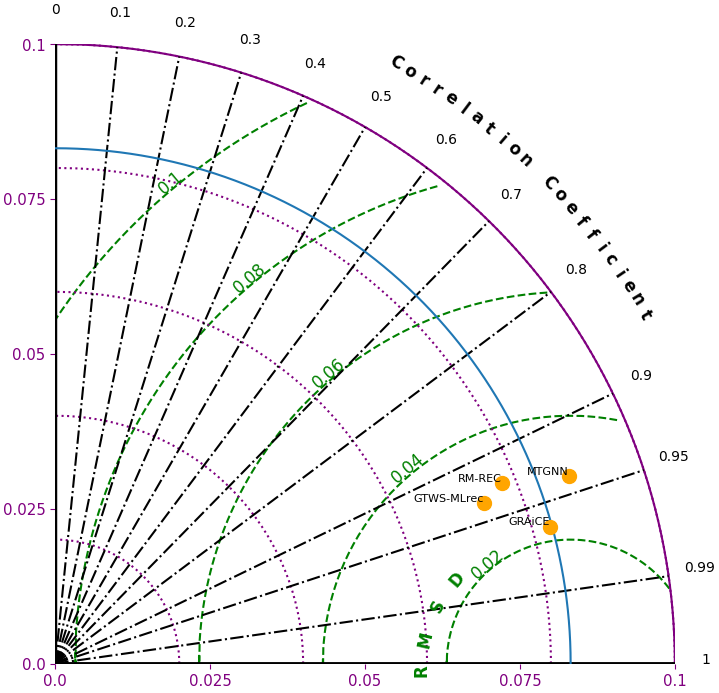}
  \caption{Taylor diagram of the reconstructions relative to GRACE, with radial axes indicating standard deviation.}
    \Description{Enjoying the baseball game from the third-base
  seats. Ichiro Suzuki preparing to bat.}
  \label{fig:taylor}
\end{figure}

The results of all reconstruction approaches correlate highly with GRACE ($>0.93$) and lie within an
RMSE band of 0.026--0.031, i.e., a few centimeters of water column. GRAiCE
performs best overall and matches the GRACE variability almost exactly;
GTWS-MLrec slightly dampens variability (std.\ 0.074), whereas MTGNN slightly
overemphasizes it (0.088). At the single Amazon cell, MTGNN is weakest
(0.866), suggesting that the graph model trades some fine-scale robustness
for its large-scale structure. Both the single-cell correlation (0.866) and the basin-mean correlation (0.939) differ from the all-cells correlation of 0.69 reported in Table~\ref{tab:quality}. The all-cells metric includes every grid cell in the domain, including regions in the arid south where correlations are relatively low. In contrast, the single-cell value represents one location in the Amazon with strong correlation, while the basin-mean value is computed from spatially averaged time series. The Taylor diagram in Figure~\ref{fig:taylor} condenses these relationships.

These numbers must be read against the predictor budget each model consumes,
shown in the \emph{Inputs} column of Table~\ref{tab:compmetrics}. While MTGNN is
driven by only precipitation, evapotranspiration, and runoff, the better-scoring baselines rely on
substantially richer feature sets. GRAiCE uses five meteorological variables
(total precipitation, snow depth water equivalent, surface net solar radiation,
surface air temperature, and relative humidity), adding solar-induced
fluorescence in its full variant \citep{palazzoli2025}. RM-REC combines
precipitation, temperature, sea surface temperature, soil moisture, runoff, and
evaporation with 17 climate indices \citep{li2021}. GTWS-MLrec ingests four whole categories of
meteorological, hydrological, land-use, and vegetation predictors
(16 variables in total) with per-cell feature selection \citep{yin2023} while DeepRec combines 16 ERA5 variables with sea surface temperature, an ENSO index, land-use and lake fractions, and engineered geographic and temporal features \citep{gentner2025}. Seen this way, the graph model
reaches a basin-mean correlation of 0.939, within 0.025 of GRAiCE and
statistically indistinguishable from GTWS-MLrec, using roughly half to a
tenth of the inputs that the better-scoring baselines require. The slight performance gap is therefore better
interpreted as a favorable accuracy-per-variable trade-off than as a deficit. Furthermore,
the topology of the spatio-temporal graph appears to substitute for part of
the information that competing models must supply through additional
covariates. This parsimony is not merely aesthetic. It lowers data-acquisition
and preprocessing cost, reduces exposure to predictors that are themselves
poorly constrained in the pre-satellite era, and makes the learned mapping
easier to attribute to physical drivers, which is an advantage we expand on in
Section~\ref{sec:extensions}.

\begin{figure}[htbp]
  \centering
  \includegraphics[width=\linewidth]{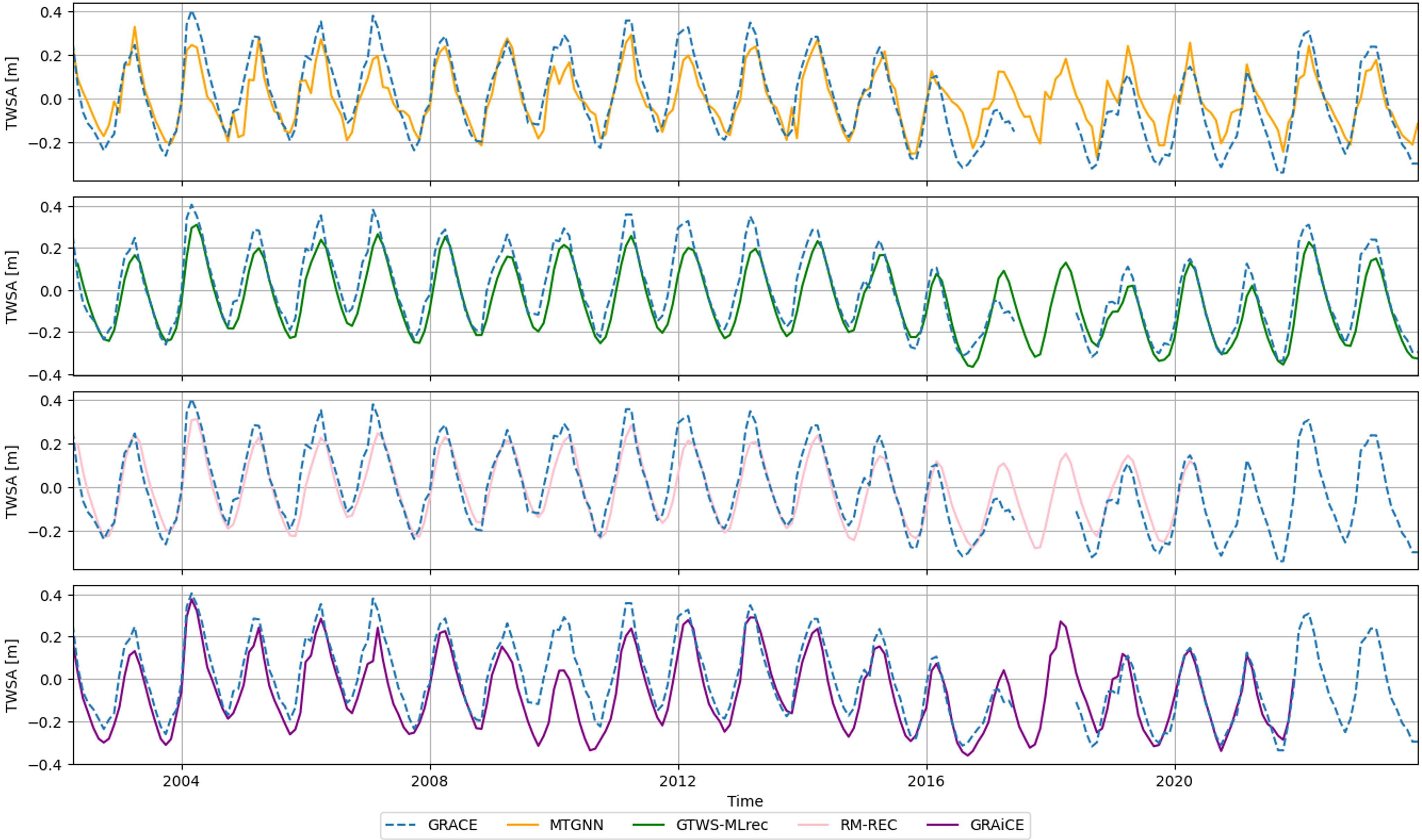}
  \caption{TWSA at the exemplary Amazon grid cell for each reconstruction approach
           vs.\ GRACE (2002--2023). The dashed blue line denotes GRACE, the orange line denotes MTGNN (ours), the green line denotes GTWS-MLrec, the pink line denotes RM-REC, and the purple line denotes GRAiCE.}
             \Description{Enjoying the baseball game from the third-base
  seats. Ichiro Suzuki preparing to bat.}
  \label{fig:temporal}
\end{figure}

The temporal view at the Amazon cell (Figure~\ref{fig:temporal}) adds nuance:
MTGNN matches the phase of the GRACE signal well---better than GTWS-MLrec,
whose phase is slightly delayed---while underestimating peak amplitudes in
parts of the record and producing a less smoothed series overall, indicating
higher sensitivity to short-term events (a potential advantage) at the cost of
possible model noise. GRAiCE, despite its leading global metrics, follows the
reference poorly at this particular cell in the first half of the record.

\begin{figure}[htbp]
  \centering
  \includegraphics[width=\linewidth]{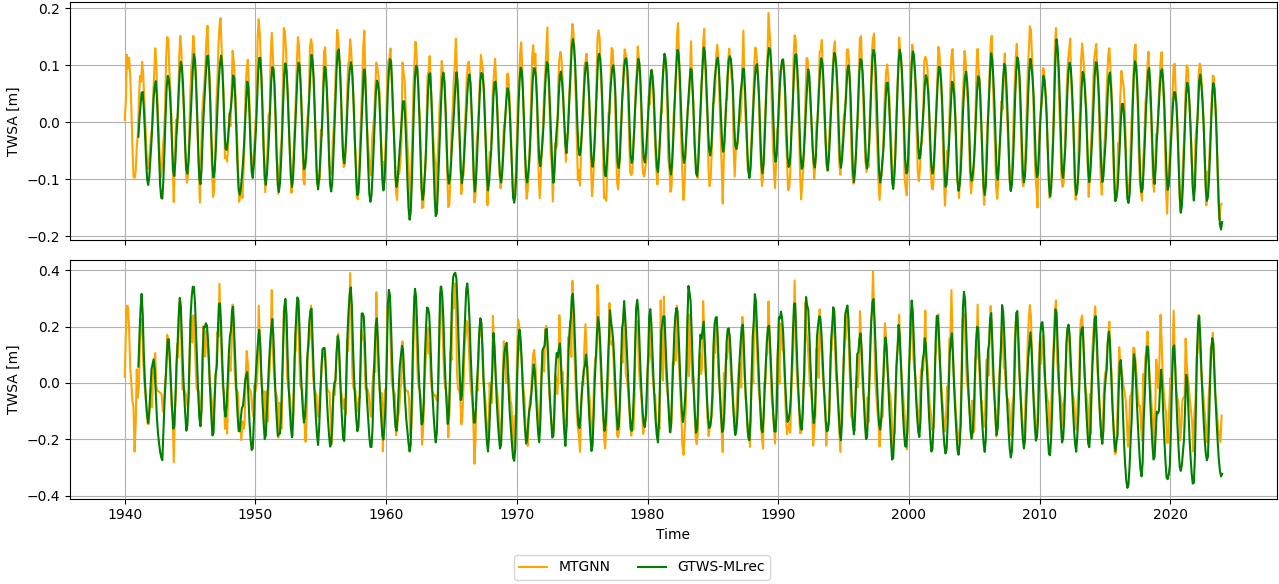}
  \caption{MTGNN (orange line) vs.\ GTWS-MLrec (green line) from 1940 onward (top: spatial mean; bottom:
           exemplary Amazon grid cell). }
             \Description{Enjoying the baseball game from the third-base
  seats. Ichiro Suzuki preparing to bat.}
  \label{fig:mtgnn-gtws}
\end{figure}

Because GTWS-MLrec is the only baseline extending back to the 1940s, it
enables a pre-GRACE consistency check (Figure~\ref{fig:mtgnn-gtws}). In the
spatial mean, MTGNN consistently shows higher amplitude but tracks the
temporal evolution closely: both products reproduce the 1948/49 decline, the
prolonged late-1960s decrease and recovery, and the decline since 2010. At the
exemplary cell, deviations alternate without systematic bias, and the major
historical excursions (1948/49, 1955, 1961, early 1970s, post-2015) agree.

\begin{figure}[htbp]
  \centering
  \includegraphics[width=\linewidth]{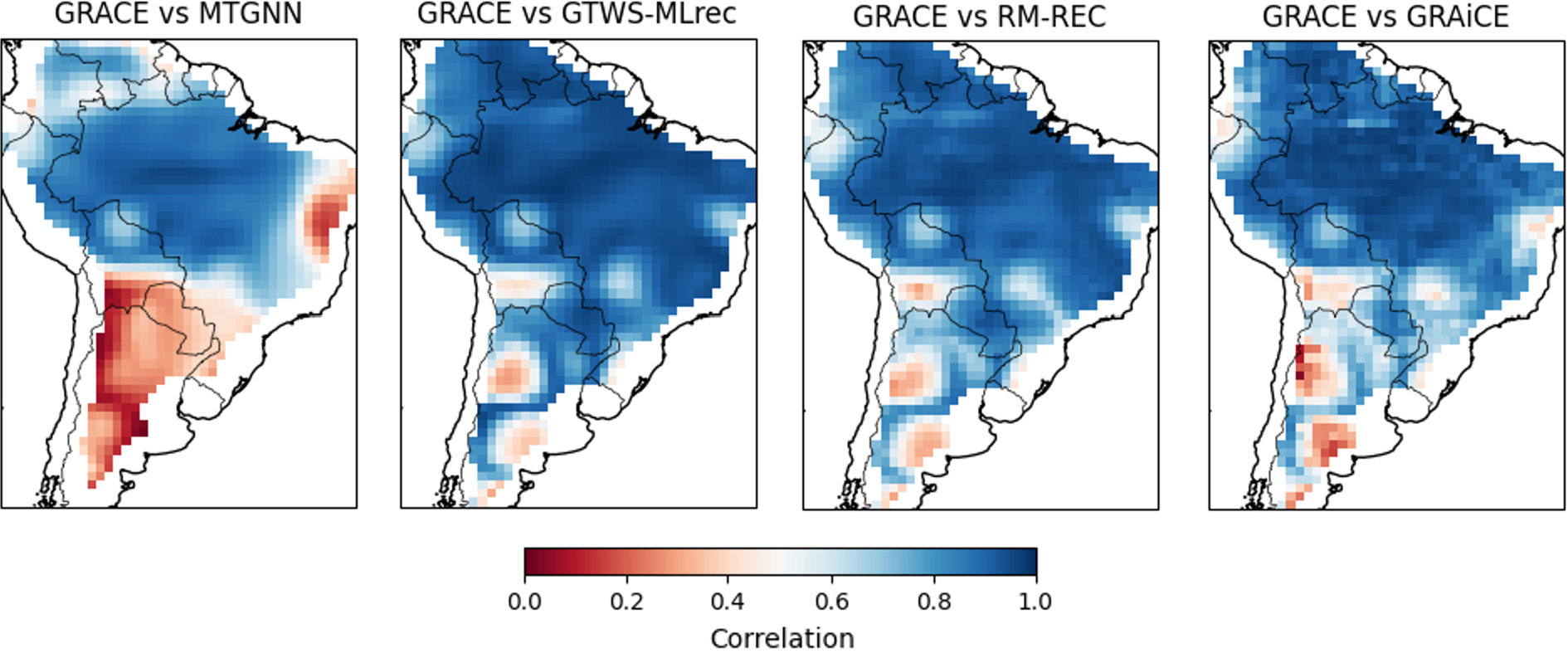}
  \caption{Spatial correlation with GRACE for all reconstruction approaches evaluated over the
           overlap period.}
             \Description{Enjoying the baseball game from the third-base
  seats. Ichiro Suzuki preparing to bat.}
  \label{fig:spatialcorr}
\end{figure}

Indicated by the spatial correlation maps (Figure~\ref{fig:spatialcorr}) 
MTGNN matches the baselines in the humid Amazon but falls
below 0.5 in parts of Bolivia, Paraguay, Argentina, and easternmost
Brazil which can be characterized as arid and semi-arid regions where hydroclimatic processes are more
complex and human influence (e.g., groundwater extraction) is stronger. The
same regions are also the weakest for the baselines, but less severely so.
This community-wide pattern echoes the limitations reported by
\citet{palazzoli2025} and \citet{yin2023} for arid, human-influenced areas.

In summary, the graph-based reconstruction is statistically on par with
established reconstruction approaches at basin scale (correlation 0.94, RMSE 0.030), captures
phase and interannual dynamics particularly well, but exhibits specific
spatial deficits in arid regions and slightly overemphasized variability.

\section{Flexible Extensions and Lessons Learned}
\label{sec:extensions}

\subsection{Input Parameter Expansion}
\label{sec:paramexp}
The reconstruction uses only the three water balance core variables, whereas
the strongest baselines integrate broader predictor sets spanning
meteorological drivers, hydrological states, and land surface/anthropogenic
factors \citep{gentner2025,yin2023,palazzoli2025,li2021}. As established in
Section~\ref{sec:comparison}, the graph model nearly matches these established approaches on
a fraction of their inputs. This raises an obvious question: if so much is
achieved with three variables, how much further could the model go---and, more importantly, how
cheaply can the additional variables be added?

The architecture makes such expansion structurally straightforward. Because each
predictor enters as an additional channel of the per-node feature dimension
$D$ (Section~\ref{sec:problem}), adding a meteorological forcing means
extending the node feature tensor from
$\mathbb{R}^{T \times N \times D}$ to $\mathbb{R}^{T \times N \times (D+1)}$
and widening the first layer. The graph topology, the temporal modules, and
the entire training pipeline are untouched. No new model has to be trained per
variable, and no per-cell feature-selection stage is required. This contrasts
with the comparing approaches. While GTWS-MLrec must rerun its ensemble and per-pixel
selection over eight input schemes when its predictor set changes
\citep{yin2023}, the per-grid-cell (Bi)LSTMs of GRAiCE are configured and
tuned cell by cell \citep{palazzoli2025}, and the patch-based CNN of DeepRec
fixes its feature stack in the patch construction \citep{gentner2025}. In the
spatio-temporal graph, by contrast, a new forcing variable is simply one more
signal observed at every node, propagated by the same shared convolutional
weights, meaning the marginal engineering cost of testing a candidate predictor is
close to zero.

As a pilot exploiting exactly this property, we appended 2\,m
temperature, known to be a key driver of evaporation, snowmelt, and soil water dynamics
a single feature channel that is consistently available in ERA5. The effect is measurable (Table~\ref{tab:temp}), reducing validation errors and increasing correlation with GRACE from 0.693 to 0.713 (grid-cell) and from 0.945 to 0.947 (basin level). Given that the architecture accepts arbitrary
numbers of node features at this near-zero cost, soil moisture and snow water
equivalent are natural next candidates, since empirical studies link soil moisture
anomalies tightly to TWS variations in semi-arid regions \citep{guo2023}. Furthermore, 
anthropogenic predictors are essential where groundwater depletion dominates
the signal \citep{asoka2020} directly targeting the arid-region weakness identified
above. The favorable accuracy-per-variable position observed in
Table~\ref{tab:compmetrics} thus reflects headroom, not a ceiling. As a result, the cheap
expansion path is precisely the one that can close the remaining gap to the
input-heavy baselines while keeping the model interpretable.

\begin{table}[htbp]
  \centering
  \caption{Validation quality without and with temperature as input.}
  \label{tab:temp}
  \begin{tabular}{l c c c c}
    \toprule
    & MSE & RMSE & MAE & Corr. \\
    \midrule
    $P,E,R$            & 0.0117 & 0.1083 & 0.0830 & 0.693 \\
    $P,E,R,T$          & 0.0105 & 0.1025 & 0.0789 & 0.713 \\
    \bottomrule
  \end{tabular}
\end{table}

\subsection{Physics-Informed Training}
\begin{figure}[htbp]
  \centering
  \includegraphics[width=1\linewidth]{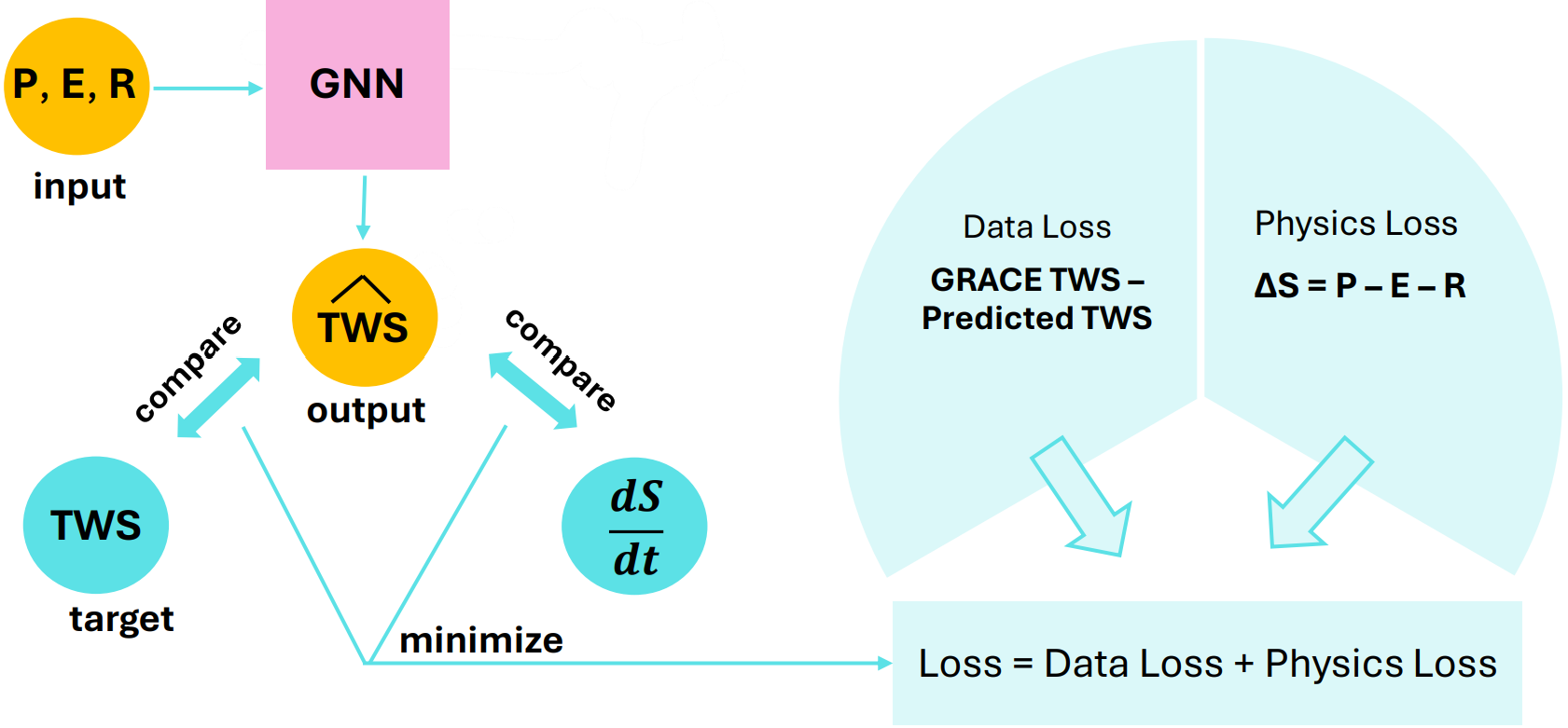}
  \caption{Hybrid framework combining the data loss with a water balance
           physics loss.}
\Description{Enjoying the baseball game from the third-base
  seats. Ichiro Suzuki preparing to bat.}
  \label{fig:pinn}
\end{figure}

Physics-informed neural networks (PINNs) integrate physical laws as soft
constraints into the loss function, improving plausibility and
generalization where data are sparse \citep{luo2025}. The water balance
(Eq.~\ref{eq:waterbalance}) translates directly into such a constraint. With
$\mathrm{TWSA}_\theta$ the network prediction, the combined loss
$L = \alpha L_{\mathrm{data}} + \beta L_{\mathrm{physics}}$ uses
\begin{equation}
\begin{split}
  L_{\mathrm{physics}} = \frac{1}{N}\sum_{i=1}^{N}
  \Bigl( & \bigl(\mathrm{TWSA}_\theta(t_i) - \mathrm{TWSA}_\theta(t_{i-1})\bigr) \\
       & - \bigl(\bar{P}(t_i) - \bar{E}(t_i) - \bar{R}(t_i)\bigr) \Bigr)^2,
\end{split}
  \label{eq:physicsloss}
\end{equation}
penalizing violations of the discrete storage recursion
$S(t) = S(t-1) + P - E - R$. Whereas \citet{gentner2025} used the water
balance for post-hoc validation, embedding it into training
(Figure~\ref{fig:pinn}) turns it into a regularizer, particularly promising
for arid regions and extreme events where purely data-driven training has
shown deficits.

This constraint is the natural consequence of the input choice. 
Because the three node features ($P$, $E$, $R$) are exactly the flux
terms of the water balance and the target ($\mathrm{TWSA}$) is its storage
term, inputs and output already stand in a closed physical relationship
(Eq.~\ref{eq:waterbalance}). The physics loss asks the network to
respect, between consecutive months, the same equation that motivated the
feature set in the first place. As a result, no auxiliary variables, derived quantities,
or separate physical model are needed to evaluate it. This represents a structural advantage of the parsimonious design over input-heavy baselines. Their larger predictor sets, including vegetation indices, land-use fractions, and radiation balances, do not naturally conform to a conservation law, making comparable constraints difficult to enforce. Here the loss is, by construction, 
hydrologically interpretable which both eases tuning of the weight
$\beta$ and makes residual violations diagnosable as genuine hydrological
inconsistencies rather than opaque model error. In short, choosing the minimal
water-balance feature set buys not only data efficiency but also a loss
function that is naturally aligned with the governing physics.

\section{Conclusion}
\label{sec:conclusion}
In this work, we applied spatio-temporal graph deep learning to reconstruct terrestrial water storage prior to the GRACE mission, addressing a long-standing problem in hydrology and satellite geodesy. An MTGNN
with a static, interpretable hybrid graph, combining geodesic proximity and
lagged climatic correlations, was trained on daily ERA5 fluxes against
monthly GRACE TWSA and used to reconstruct South American water storage
anomalies back to 1940. Against GRACE/GRACE-FO the reconstruction achieves a
grid-cell correlation of 0.69, a basin-mean correlation of 0.94 with near-zero 
bias, and correctly reproduces the spatial
fingerprints of major ENSO events. Compared with three established reconstruction methods (GTWS-MLrec, RM-REC, and GRAiCE), our approach achieves competitive basin-scale performance using only three predictor variables, while sharing the common challenge of reduced accuracy in arid and human-influenced regions.


This case study demonstrates that graph architectures matured on urban mobility problems transfer to global
geophysical fields once the graph encodes domain structure such as teleconnections. Our results show that an analogy between traffic flow in road networks and water transport in the climate system is not merely conceptual but can be directly exploited in graph-based architectures. They also highlight that the choice between learned and prescribed graphs is a key modeling decision rather than an implementation detail.
For the field of hydrology and geodesy, the resulting reconstruction offers a new,
topology-aware perspective on historical water storage that complements
existing per-cell reconstruction approaches. Ongoing work scales the workflow from the South
American test region to the global $1^\circ$ grid, expands the predictor set
toward hydrological states and anthropogenic factors, and integrates the
water balance constraint into training.


\bibliographystyle{ACM-Reference-Format}
\bibliography{software}

\end{document}